\pdfoutput=1

\documentclass[11pt]{article}

\usepackage{EMNLP2023}

\usepackage{times}
\usepackage{latexsym}

\usepackage[T1]{fontenc}

\usepackage[utf8]{inputenc}

\usepackage{microtype}

\usepackage{inconsolata}

\title{AMR Parsing with Instruction Fine-tuned Pre-trained Language Models}

\author{
Young-Suk Lee,  Ramón Fernandez Astudillo, Radu Florian, Tahira Naseem, Salim Roukos \\
 \texttt{\{ysuklee,raduf,tnaseem,roukos\}@us.ibm.com} \\
 \texttt{ramon.astudillo@ibm.com} \\ 
 IBM Research AI
  }

\begin{document}
\maketitle
\begin{abstract}
Instruction fine-tuned language models on a collection of instruction annotated datasets (FLAN) have shown highly effective to improve model performance and generalization to unseen tasks. However, a majority of standard parsing tasks including abstract meaning representation (AMR), universal dependency (UD), semantic role labeling (SRL) has been excluded from the FLAN collections for both model training and evaluations. In this paper, we take one of such instruction fine-tuned pre-trained language models, i.e. FLAN-T5, and fine-tune them for AMR parsing. Our extensive experiments on various AMR parsing tasks including AMR2.0, AMR3.0 and BioAMR indicate that FLAN-T5 fine-tuned models out-perform previous state-of-the-art models across all tasks. In addition, full fine-tuning followed by the parameter efficient finetuning, LoRA, further improves the model performances, setting new state-of-the-arts in Smatch on AMR2.0 (86.4), AMR3.0 (84.9) and BioAMR (82.3).

\end{abstract}

\section{Introduction}

Instruction fine-tuning language models on a collection of annotated datasets has proven highly effective to improve model performance and generalization to unseen tasks both in general purpose open domain setup, as in \cite{flan-t5-2022, flan2021,flan2023,instructgpt, naturalinstructions, supernaturalinstructions, unnatural2022} and specialized tasks such as conversational dialogs in \cite{instructdial2022}.

Despite its great success in the majority of natural language processing tasks, however, standard parsing tasks such as abstract meaning representation (AMR), \cite{banarescu-etal-2013-abstract, bevilacqua2021aaai,zhou2021emnlp, bai-etal-2022-graph}, universal dependency (UD), \cite{ud2017}, semantic role labeling (SRL), \cite{srl-gildea2002, propbank2005}, etc. have been largely excluded from the fine-tuned language net (FLAN) collections either for model training or evaluations. And therefore it still remains to be seen whether or not instruction fine-tuned language models are as effective for standard parsing tasks as for other NLP tasks.

In this paper, we fine-tune FLAN-T5 models of \citet{flan-t5-2022} (FLAN\-T5\-Large and FLAN\-T5\-XL) on a wide range of AMR parsing tasks including AMR2.0, AMR3.0 and BioAMR. We show that fine-tuning FLAN-T5 models on AMR parsing leads to a significant improvement over the previous BART fine-tuned SoTA models by \citet{zhou2021emnlp, bai-etal-2022-graph}. We further explore a parameter efficient fine-tuning technique, LoRA (Low Rank Adaptation), \cite{lora2021}. While LoRA\-only fine\-tuned models do not out-perform full fine-tuned models, full fine-tuning followed by LoRA fine-tuning significantly improve full fine-tuned models, setting new state-of-the-arts across all AMR parsing tasks.

Our main contributions are as follows:

\begin{itemize}
    \item We apply instruction fine-tuned FLAN-T5 models to AMR parsing for the first time. We show that FLAN-T5 fine-tuned AMR parsing models significantly out-perform previous BART fine-tuned SoTA models.
    \item We explore the parameter efficient fine-tuning technique LoRA for sequence-to-sequence tasks. Although fine-tuning FLAN-T5 models with LoRA only does not out-perform full fine-tuned models, LoRA fine-tuning of full fine-tuned models further improves model performances.
    \item We push the envelope of AMR parsing, by setting new SoTA in Smatch on AMR2.0 (86.4), AMR3.0 (84.9) and BioAMR (82.3).
\end{itemize}

\section{AMR Parsing with FLAN-T5 Models}


Flan-T5 models, \cite{flan-t5-2022}, are obtained by instruction fine-tuning T5-LM adapted models, \cite{prompttuning2021}, on a collection of 1.8K instruction annotated tasks.\footnote{https://github.com/google-research/FLAN/blob/main/flan/v2/flan\_collection\_info.csv}   
They are prefix language models\footnote{Given natural text prefix as input, the model must produce the natural text continuation as output.} and achieve strong few-shot performance even compared to much larger models, such as PaLM 62B. 

Like all models derived from T5 models, \cite{t5-raffel}, we pose AMR parsing as a text-to-text problem and train models to transfer a text to a linearized AMR graph with the task prefix \textbf{amr generation}. FLAN-T5 model size variants, all of which use 32,128 unique vocabulary, is shown in Table~\ref{tab:flan-t5-model-sizes}. 

\begin{table*}[!tb]
\centering
\begin{tabular}{l|cccccc}
\hline
Models & \# parameters & \# layers & d$_{model}$ & d$_{ff}$ & d$_{kv}$ & \# heads \\\hline
Flan-T5-Small  & 77M & 8 & 512 & 1024 & 64 & 6 \\
Flan-T5-Base & 250M & 12 & 768 & 2048 & 64 & 12 \\
\textbf{Flan-T5-Large} & 780M & 24 & 1024 & 2816 & 64 & 16 \\
\textbf{Flan-T5-XL} & 3B & 24 & 2048 & 5120 & 64 & 32 \\
Flan-T5-XXL & 11B & 24 & 4096 & 10240 & 64 & 64 \\\hline
\end{tabular}
\caption{Flan-T5 model size variants obtained from each model configuration file of https://huggingface.co/models. We fine-tune Flan-T5-Large and Flan-T5-XL for our AMR experiments.}
\label{tab:flan-t5-model-sizes}
\end{table*}

\subsection{Pre- and Post-processing}

We first remove wiki tags from the raw AMR graphs. We then serialize the AMR graph and transfer the node variable information of concepts to the concepts themselves.\footnote{https://github.com/IBM/graph\_ensemble\_learning} If a graph includes the same concept more than once, unique indices are appended to each concept, e.g. \textit{thing\_1, thing\_2}, etc. Finally, we add the task prefix \textbf{amr generation} to each input text. A sample input text and the corresponding serialized AMR graph after pre-processing is shown in Figure~\ref{fig:preprocess}.

\begin{figure}[ht]
\small
\begin{verbatim} 
Input: Statistics also revealed that
Taiwanese business investments in the 
mainland is tending to increase

AMR graph:
(r / reveal-01
    :ARG0 (s / statistic)
    :ARG1 (t / tend-02
        :ARG1 (t2 / thing
            :ARG1-of (i / invest-01
            :ARG0 (c / country
                :wiki "Taiwan"
                :name (n / name
                      :op1 "Taiwan"))
            :ARG2 (m / mainland)
            :mod (b / business)))
    :ARG2 (i2 / increase-01
            :ARG1 t2))
    :mod (a / also))
    
Serialized AMR graph:
( reveal-01 :ARG0 ( statistic ) :ARG1 
( tend-02 :ARG1 ( thing :ARG1-of ( invest-01 
:ARG0 ( country :name ( name :op1 "Taiwan" ) )
:ARG2 ( mainland ) :mod ( business ) ) ) 
:ARG2 ( increase-01 :ARG1 thing ) ) :mod 
( also ) )

Input text with the task prefix:
amr generation ; Statistics also revealed 
that Taiwanese business investments in the 
mainland is tending to increase
\end{verbatim}

\caption{Serialization of AMR graphs and addition of the task prefix \textbf{amr generation} to the input text. In the serialized graph, the variables corresponding to each concept in the original AMR graph is removed while the parentheses indicating the concept spans are retained. The input text with the task prefix and the serialized AMR graphs are used for model training.}
\label{fig:preprocess}
\end{figure}

For testing, the decoder first generates serialized graphs, which is then de-serialized to restore the concept variables including re-ifications. We finally restore wiki tags using deterministic algorithms. 

\subsection{Full Fine-tuning}

For full fine-tuning, we call the huggingface transformers\footnote{https://github.com/huggingface/transformers} class \texttt{T5ForConditionalGeneration} and \texttt{T5Tokenizer}. We train all models on 2 NVIDIA A100 80GB machines for 24 hours.

For both FLAN-T5-large and FLAN-T5-XL models, we set the maximum source and target lengths to 512. Learning rate is set to 5e-5. We utilize sentence based batching for mini batches. Batch size is 8 for FLAN-T5-large and 4 FLAN-T5-XL, distributed over 2 GPUs. 

We run the validation data set after each epoch and choose the model with the highest validation set Smatch score as the final best model for testing.

\subsection{LoRA Fine-tuning}

We experiment with the low rank adaptation LoRA\footnote{https://github.com/huggingface/peft} for AMR parsing, a sequence-to-sequenc task using an encoder-decoder architecture.

Largely following the recommended setup of adapting only the \textit{q} (query) and \textit{v} (value) projections in the transformer,  we explore the LoRA configurations  between \texttt{rank=8, alpha=32} and \texttt{rank=16, alpha=64} while fixing \texttt{task\_type} to \texttt{SEQ\_2\_SEQ\_LM}. We call model.eval() to merge LoRA parameters with the corresponding pre-trained ones after each parameter update and for inferencing and model.train()  to split the LoRA parameters from the pre-trained ones for updating only LoRA parameters. Unlike full fine-tuning for which learning rate is 5e-5, we use learning rate 4e-1 for LoRA fine-tuning. We have found that LoRA fine-tuning requires a higher learning rate than full fine-tuning for optimal performances, which is not surprising given the much fewer number of parameters to update with LoRA fine-tuning compared with full fine-tuning.

We experiment with two different modes of LoRA fine-tuning. First, apply LoRA fine-tuning to the pretrained language model (PLM) directly. Second, apply full fine-tuning to the PLM, and then apply LoRA fine-tuning to the full fine-tuned models. LoRA fine-tuning on PLM directly does not seem to improve the performances over full fine-tuned models.\footnote{We have not explored the full range of adjustable parameters of LoRA, e.g. \textit{k} and \textit{o} projections in addition to \textit{q} and \textit{v} projections, or values other than 8 and 16 for LoRA rank and 32 and 64 for LoRA alpha.} However, LoRA fine-tuning on full fine-tuned models improve the full fine-tuned models across various AMR parsing tasks and model configurations.

\section{Experimental Results}

We experiment on 3 AMR tasks, AMR2.0, AMR3.0 and BioAMR, and 2 model training configurations, FLAN-T5-large and FLAN-T5-XL.
Training and test corpora statistics are shown in Table~\ref{tab:benchmarkdata}. Silver training corpus is annotated with the MBSE ensemble distillation technique presented in \cite{mbse2022naacl}. 

\begin{table}[!t]
\centering
\begin{tabular}{l|l|rr}
\hline
\hline
\textbf{Dataset}  & \textbf{Split} & \textbf{Sents} & \textbf{Tokens}  \\
\hline 
AMR2.0   & Train$^{h}$ & 36,521 & 653K \\
        & Test   & 1,371 & 30K    \\\hline
AMR3.0   & Train$^{h}$  & 55,635 & 1M  \\ 
        & Test   & 1,898 & 39K   \\\hline 
Bio AMR  & Train$^{h}$  & 5,452 & 231K \\
         & Test   & 500 & 22K  \\\hline
PropBank   & Silver$^{std}$ & 20K & 386K  \\
SQuAD2.0-C    & Silver$^{std}$ & 70K & 2M  \\
Ontonotes5.0  & Silver$^{std}$ & 59K & 1.1M  \\
WikiText-103  & Silver$^{std}$ & 70K & 2M \\\hline 
BioNLP-ST-2011  & Silver$^{bio}$ & 15K & 460K \\
CRAFT  & Silver$^{bio}$ & 27K & 740K \\
PubMed & Silver$^{bio}$ & 26K & 750K \\\hline

\end{tabular}
\caption{Corpus statistics for the standard benchmark experiments on AMR2.0, AMR3.0 and BioAMR test sets. In corpus split, Train$^{h}$ indicates human annotated treebank. Silver$^{std}$ indicates the unlabeled data for silver training of AMR2.0 and AMR3.0. Silver$^{bio}$ indicates the unlabeled data for silver training of BioAMR. Silver training corpus is annotated with the MBSE ensemble distillation technique in \cite{mbse2022naacl}.}
\label{tab:benchmarkdata}
\end{table}

\begin{table*}[t]
\centering
\begin{tabular}{l|ccc||ccc}
\hline
\textbf{Training Corpora}  & \multicolumn{3}{c||}{\textbf{Human Annotations}} & \multicolumn{3}{c}{\textbf{Human \& Silver Annotations}} \\
\hline
\textbf{Models}  & \textbf{AMR2.0} & \textbf{AMR3.0} & \textbf{BioAMR}  & \textbf{AMR2.0} & \textbf{AMR3.0} & \textbf{BioAMR}\\\hline 
\cite{mbse2022naacl} & 84.2 & 82.3 & 79.8 & 85.9 & 84.3 & 81.3 \\\hline
FLAN-T5-Large-LoRA & 82.3 & 81.7 & 79.1 & 84.6 & 83.0 & 80.4 \\
FLAN-T5-Large-FFT    & 84.6 & 83.2 & 81.0 & 85.8 & 84.6 & 82.1 \\
FLAN-T5-Large-FFT-LoRA & 84.8{\small{$\pm$0.1}} & 83.3{\small{$\pm$0.0}} & 81.2{\small{$\pm$0.1}} & 86.1{\small{$\pm$0.0}} & 84.7{\small{$\pm$0.1}} & 82.2{\small{$\pm$0.1}} \\\hline
FLAN-T5-XL-FFT      & 84.6 & 83.4 & 80.9 & 86.1 & 84.6 & 82.3 \\
FLAN-T5-XL-FFT-LoRA & 84.8{\small{$\pm$0.0}} & 83.9{\small{$\pm$0.1}} & 81.6{\small{$\pm$0.0}}  & 86.4{\small{$\pm$0.1}} & 84.9{\small{$\pm$0.0}} & 82.3{\small{$\pm$0.2}} \\
\hline
\end{tabular}
\caption{Performance of FLAN-T5 fine-tuned models trained on human annotations only (left) and human and silver annotations (right). FFT denotes full fine-tuning. The numbers prefixed by $\pm$ indicate the standard deviation of Smatch scores across 2 seeds with different LoRA configurations.}
\label{tab:flan-t5-fft-lora}
\end{table*}

\begin{table*}
\centering
\begin{tabular}{l|c|c|c|c|c}
\hline
\textbf{Models} & \textbf{PLM} & \textbf{Silver} & \textbf{AMR2.0} & \textbf{AMR3.0} & \textbf{BioAMR} \\\hline
SPRING \cite{bevilacqua2021aaai} & BART-large & - & 84.5 & 83.0 & 79.9 \\
SPRING \cite{bevilacqua2021aaai} & BART-large & 200K & 84.3 & 83.0 & 59.5 \\
StructBART-vanilla \cite{zhou2021emnlp} & BART-large & 90K & 84.7 & 82.7 &  - \\
BARTAMR \cite{bai-etal-2022-graph} & BART-large & 200K & 85.4 & 84.2 & 63.2 \\
StructBART-MBSE \cite{mbse2022naacl} & BART-large & 219K & 85.9 & 84.3 & 81.3 \\
FLAN-T5-Large-FFT-LoRA (Ours) & FLAN-T5-Large & 219K & 86.1 & 84.7$\ddagger$ & 82.2$\ddagger$ \\
FLAN-T5-XL-FFT-LoRA (Ours) & FLAN-T5-XL & 219K & \textbf{86.4}$\ddagger$ & \textbf{84.9}$\ddagger$ & \textbf{82.3}$\ddagger$ \\\hline
\end{tabular}
\caption{Comparison of AMR parsing models for AMR2.0, AMR3.0 and BioAMR test sets. We compare the current FLAN-T5 fine-tuned models (Ours) against those BART-large fine-tuned models. Boldface indicates the best model scores. $\ddagger$ indicates that the model is statistically significantly better than all of the previous models at p=0.05 according to randomized bootstrap statistical significance tests.}
\label{tab:compareAMRs}
\end{table*}

\begin{table*}
    \centering
    \begin{tabular}{l|c|c|c|c|c|c|c|c|c} \hline
Test Data & Smatch & Unlabel & NoWSD & Concepts & NER & Neg. & Wiki & Reentrancy & SRL \\\hline 
AMR2.0-mbse  & 86.4 & 89.2 & 86.7 & 92.0 & 93.0 & 79.0  & 81.0 & 78.0 & 85.0 \\
AMR3.0-mbse & 84.9 & 87.8 & 85.3  & 91.0  & 90.0 & 75.0  & 78.0 & 77.0 & 84.0 \\
BioAMR-mbse & 82.5 & 85.1 & 82.4  & 89.0  & 81.0  & 80.0 & 0.0  & 74.0 & 83.0 \\\hline
AMR2.0-base & 84.8 & 87.8 & 85.2 & 91.0 & 93.0 & 75.0 & 78.0 & 76.0 & 83.0 \\
AMR3.0-base & 83.9 & 86.8 & 84.3 & 90.0 & 89.0 & 74.0 & 72.0 & 76.0 & 83.0 \\
BioAMR-base & 81.6 & 84.1 & 81.5 & 88.0 & 82.0 & 80.0 & 0.0 & 72.0 & 81.0 \\\hline
    \end{tabular}
    \caption{Fine-grained F1 scores of FLAN-T5-XL-FFT-LoRA models on test data sets. AMR2.0-mbse, AMR3.0-mbse, BioAMR-mbse indicate AMR2.0, AMR3.0 and BioAMR test sets are evaluated on models trained with the combination of human annotated and silver training data discussed in \cite{mbse2022naacl}. AMR2.0-base, AMR3.0-base and BioAMR-base indicate AMR2.0, AMR3.0 and BioAMR test sets are evaluated on models trained with human annotated data only.}
    \label{tab:finegrainedscore}
\end{table*}

Experimental results are shown in Table~\ref{tab:flan-t5-fft-lora}. As a point of reference, we include the highest Structured BART scores from \citet{mbse2022naacl}. FFT denotes full fine-tuning and LoRA, LoRA fine-tuning. All FLAN-T5-Large-FFT-LoRA and FLAN-T5-Large-FFT-LoRA scores are an average of two distinct model scores, one trained with \texttt{lora\_rank=8, lora\_alpha=32} and the other, with \texttt{lora\_rank=16, lora\_alpha=64}. 

Across all tasks and training corpus sizes, full fine-tuned (FFT) FLAN-T5-Large models out-perform Structured BART, except for AMR2.0 with silver training data for which FLAN-T5-Large is 0.1 Smatch lower than Structured BART. Full fine-tuned FLAN-T5-XL models out-perform all corresponding Structured BART models.
LoRA fine-tuning lags behind full fine-tuning in performance when comparing FLAN-T5-Large-LoRA with FLAN-T5-Large-FFT. Full fine-tuning followed by LoRA fine-tuning (FFT-LoRA), however, always out-performs full fine-tuning only except for BioAMR with silver training corpus, for which full fine-tuned model score is the same as that of full fine-tuning followed by LoRA fine-tuning. 

The fact that full fine-tuning followed by LoRA fine-tuning improves the scores of full fine-tuned models is somewhat unexpected and this training setup does not seem to have been explored elsewhere, \cite{baize2023, dylora2022, peft2023}. We conjecture that the low rank adaptation by LoRA prevents the model from over-fitting the training data especially when the supervised training corpus size is large as in our AMR parsing setup, which in turn leads to better generalization capabilities on unseen test sets. We leave this topic for future research for now.

We compare the performances of our best models with previous SoTA models in Table~\ref{tab:compareAMRs}. We restrict the comparisons only with models that fine-tune PLMs, i.e. BART-large.  The model scores with $\ddagger$ indicates that the models are significantly better than all of the previous models at p=0.05 according to randomized bootstrap statistical significance tests.\footnote{https://github.com/IBM/transition-amr-parser/blob/master/scripts/smatch\_aligner.py} We see that FLAN-T5 fine-tuned models out-perform the previous SoTA models across all training conditions and model configurations.

We show the detailed scores of our best models both with and without silver training corpus in Table~\ref{tab:finegrainedscore}. BioAMR wiki scores are 0.0 because we do not apply wikification to BioAMR parsing outputs. Overall, concept scores are the highest and negation/re-entrancy are the lowest among all categories. Negation scores are lower than re-entrancy scores for AMR3.0-mbse, AMR3.0-base and AMR2.0-base whereas for all others, re-entrancy scores are lower than negation scores. Named entity detection (NER) does not seem to be an issue even with BioAMR, which should be attributed to the fact that the training corpus includes 5K human annotated BioAMR graphs.
\section{Conclusion and Future Work}

We presented AMR parsing with instruction fine-tuned FLAN-T5 models, first parsing results with FLAN-T5 models to the best of our knowledge. The experimental results indicate that
FLAN-T5 fine-tuned AMR parsing models significantly out-perform previous SoTA models, which were also fine-tuned with another PLM, BART-large. We also explore parameter efficient fine-tuning LoRA. While LoRA fine-tuned models under-perform full fine-tuned models, LoRA tuning applied to full fine-tuned models further improves the Smatch scores of fine-tuned models across all training conditions. We push the envelope of AMR parsing, by setting new SoTA in Smatch on AMR2.0 (86.4), AMR3.0 (84.9) and BioAMR (82.5), which is even higher than 7 model graphene ensemble results presented in Tables 2 and 3 of \cite{graphensemble2021}.

While full fine-tuning followed by LoRA fine-tuning improves the model performances significantly, compared with the models with full fine-tuning only, we do not yet understand exactly why this should be the case, which we leave this for future research. 

With the advent of very powerful instruction fine-tuned language models with human feedback such as ChatGPT and GPT-4, many natural language processing tasks, including classification and detection, achieve very high zero-shot performances. Nontheless, given the unique label vocabulary and the hidden structure present in most parsing representations, zero shot parsing on a new parsing task does not seem easily achievable. Instruction fine-tuning on a collection of annotated natural language parsing tasks, along the lines what has been done for dialog tasks in \citet{instructdial2022}, might lead to high performing few-shot or zero-shot learning of new parsing tasks.


\bibliography{revisedpaper,flan}
\bibliographystyle{acl_natbib}

\newpage



\end{document}